\documentclass[review]{elsarticle}

\usepackage{lineno,hyperref}
\usepackage{amsmath}
\usepackage{multirow}
\modulolinenumbers[5]

\journal{Journal of \LaTeX\ Templates}

\begin{document}

\begin{frontmatter}

\title{Semantic-diversity transfer network for generalized zero-shot learning via inner disagreement based OOD detector}

\author[mymainaddress,mysecondaryaddress]{Bo Liu}
\author[mymainaddress,mythirdaddress,myfourthaddress]{Qiulei Dong\corref{mycorrespondingauthor}}
\cortext[mycorrespondingauthor]{Corresponding author}
\ead{qldong@nlpr.ia.ac.cn}
\author[mymainaddress,mysecondaryaddress]{Zhanyi Hu}

\address[mymainaddress]{National Laboratory of Pattern Recognition, Institute of Automation, Chinese Academy of Sciences, Beijing 100190, China}
\address[mysecondaryaddress]{School of Future Technology, University of Chinese Academy of Sciences, Beijing 100049, China}
\address[mythirdaddress]{School of Artificial Intelligence, University of Chinese Academy of Sciences, Beijing 100049, China}
\address[myfourthaddress]{Center for Excellence in Brain Science and Intelligence Technology, Chinese Academy of Sciences, Shanghai 200031, China}

\begin{abstract}
Zero-shot learning (ZSL) aims to recognize objects from unseen classes, where the kernel problem is to transfer knowledge from seen classes to unseen classes by establishing appropriate mappings between visual and semantic features. The knowledge transfer in many existing works is limited mainly due to the facts that $(\romannumeral1)$ the widely used visual features are global ones but not totally consistent with semantic attributes; $(\romannumeral2)$ only one mapping is learned in existing works, which is not able to effectively model diverse visual-semantic relations; $(\romannumeral3)$ the bias problem in the generalized ZSL (GZSL) could not be effectively handled. In this paper, we propose two techniques to alleviate these limitations. Firstly, we propose a \textbf{Se}mantic-diversity \textbf{t}ransfer \textbf{Net}work (SetNet) addressing the first two limitations, where 1) a multiple-attention architecture and a diversity regularizer are proposed to learn multiple local visual features that are more consistent with semantic attributes and 2) a projector ensemble that geometrically takes diverse local features as inputs is proposed to model visual-semantic relations from diverse local perspectives. Secondly, we propose an inner disagreement based domain detection module (ID3M) for GZSL to alleviate the third limitation, which picks out unseen-class data before class-level classification. Due to the absence of unseen-class data in training stage, ID3M employs a novel self-contained training scheme and detects out unseen-class data based on a designed inner disagreement criterion. Experimental results on three public datasets demonstrate that the proposed SetNet with the explored ID3M achieves a significant improvement against $30$ state-of-the-art methods.
\end{abstract}


\end{frontmatter}


\section{Introduction}
Despite the great success of deep learning on object recognition, its superiority heavily relies on the expensive labeling resources. Recently, zero-shot learning (ZSL) has received much attention in the machine learning and computer vision fields, which aims to recognize objects from those classes without labeled samples (often named as unseen classes). To deal with the absence of unseen-class labels, the semantic features (e.g. attributes and text descriptions) which represent some semantic relations between seen and unseen classes are used as side information in ZSL. The key to achieve ZSL is to transfer the learned visual-semantic knowledge from seen to unseen classes by establishing appropriate mappings between visual features (often extracted by a convolutional neural network (CNN)) and semantic features. 

According to the mapping direction, most existing ZSL methods could be roughly divided into three categories: semantic-to-visual mapping methods, intermediate mapping methods, and visual-to-semantic mapping methods. The semantic-to-visual mapping methods~\cite{zhang2017DEM,Changpinyo17EXEM,paul2019SABR,xian2019f-VAEGAN,keshari2020OCDZSL,Vyas2020LrGAN,XuXWSA20} either map individual semantic features into individual visual prototypes via a regression model, or map individual semantic features into many visual feature instances via a conditional generative model, and then learn a classifier with the inferred unseen-class visual prototypes or visual features in the visual feature space. The intermediate mapping methods~\cite{ding2019MLSE,Changpinyo16SYNC,Meng2019ZeroShotLV,Peng2018JointSA,Tsai2017ReViSE} map both the visual features and semantic features into a common latent space and then learn a classifier in the latent space. Despite their high performances, it is not easy for the two categories of methods to integrate the feature mapping and the feature extractor in a whole network which can be trained end-to-end. The visual-to-semantic mapping methods~\cite{Frome13DeViSE,Xie_2019_AREN,Zhu19SGML,huynh2020DAZLE,Yu18SGA,Li18ldf,HuynhE20} employ a CNN to extract visual features and map them into the semantic feature space by a mapping function on the top of the CNN, and then classify them in the semantic feature space. Although the performances of the visual-to-semantic methods are probably affected by the hubness problem~\cite{zhang2017DEM}, a main advantage of these methods is that they could naturally explore a new CNN architecture and jointly learn the feature extractor and the feature mapping in an end-to-end manner, so that they could extract more semantically consistent visual features.

One open problem in ZSL is the domain shift problem that the mapping function between visual and semantic features learned with data from a seen-class domain has a limited transferability to an unseen-class domain. The limited transferability is potentially caused by $(\romannumeral1)$ the patterns of knowledge transfer from a seen-class domain to an unseen-class one are generally diverse, for instance, heads of different animals have different visual appearances, hence they should have different mapping relationships with the same `head' semantic attribute. However, most existing methods learn only one mapping function that is not able to sufficiently deal with such a diversity of knowledge transfer; $(\romannumeral2)$ the annotated semantic attributes are fine-grained ones, e.g. head and hand attributes, while the widely used visual features extracted by a pre-trained CNN in existing works are generally global ones, which are not totally consistent with semantic attributes; $(\romannumeral3)$ the recognition bias to the seen-class domain in the generalized ZSL (GZSL) could not be effectively handled.

In this paper, we propose two techniques to alleviate the above limitations. Firstly, we propose a semantic-diversity transfer network (SetNet) for ZSL to alleviate the first two limitations $(\romannumeral1)$ and $(\romannumeral2)$,
where a multiple-attention architecture is employed to extract multiple local visual features that are encouraged to be semantically diverse via a diversity regularizer, and a visual-semantic projector ensemble with population diversity is adopted to model the diverse relations between local visual features and semantic features. Different projectors in the ensemble geometrically take different local visual features as inputs and employ different projection functions to model the visual-semantic relations, which has the advantage of transferring knowledge from different local perspectives. Secondly, we propose an inner disagreement based domain detection module (ID3M) for GZSL to alleviate the third limitation $(\romannumeral3)$, which aims to detect out the unseen-class data before class-level classification. In the proposed ID3M, due to the absence of real unseen-class data, a novel self-contained training scheme which trains multiple sub-domain detection modules (sub-DDMs) with virtual unseen-class data is explored and an inner disagreement criterion is designed for differentiating seen and unseen class data with these sub-DDMs.

\begin{figure}[t]
	\centering
	\includegraphics[width=0.9\linewidth]{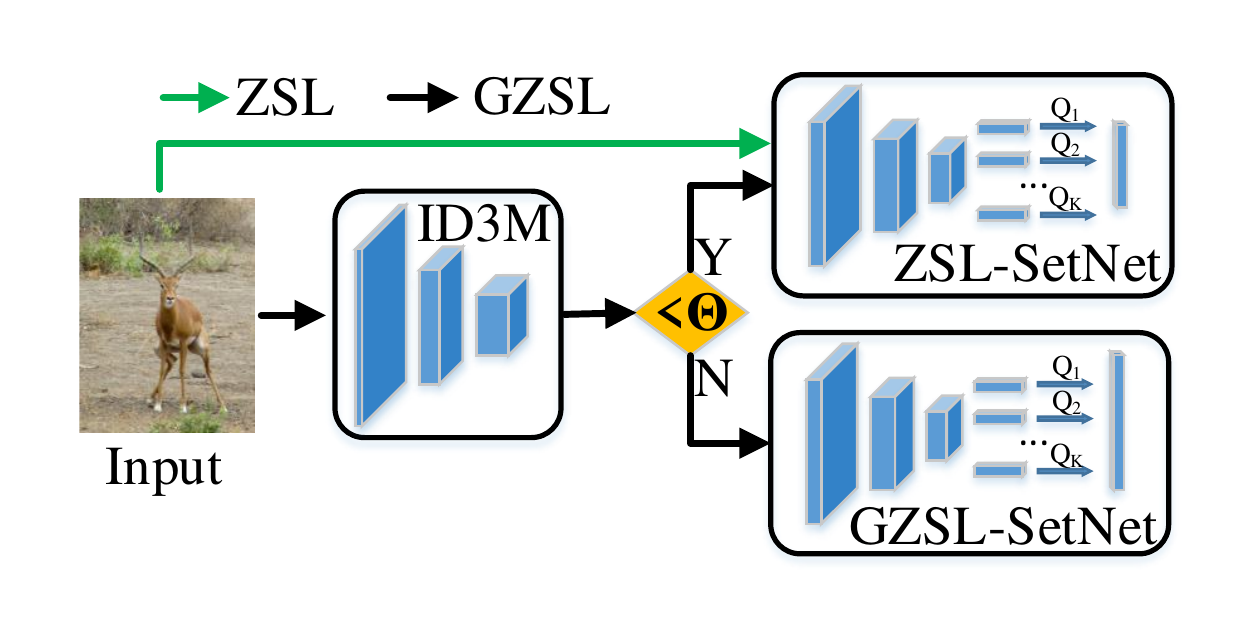}
	\caption{Flowchart of the proposed method.}
	\label{fig1}
\end{figure}

By integrating SetNet with ID3M, a novel method is further proposed to tackle the (G)ZSL problem, whose flowchart is shown in Figure~\ref{fig1}. In the flowchart, for ZSL, the inputs are directly processed by a SetNet for ZSL (i.e. ZSL-SetNet). For GZSL, the inputs are firstly fed into ID3M to perform domain detection. If the inputs are from the unseen-class domain, they will be processed by ZSL-SetNet, otherwise by a SetNet for GZSL (i.e. GZSL-SetNet). In summary, our contributions are three-fold:
\begin{itemize}
	\item We propose a semantic-diversity transfer network (SetNet) to consolidate knowledge transfer from a seen-class domain to an unseen-class one. SetNet extracts diverse local visual features which are consistent with semantic attributes via a multiple-attention module and a diversity regularizer, and establishes a visual-semantic projector ensemble which models the visual-semantic relations from diverse local perspectives.
	\item We propose an inner disagreement based domain detection module (ID3M) to alleviate the bias problem in GZSL, which employs a novel self-contained training scheme and detects out unseen-class data based on an inner disagreement criterion.
	\item Extensive experimental results demonstrate that the proposed SetNet with the explored ID3M can outperform $30$ state-of-the-art methods with significant improvements on three public datasets with two data splits.
\end{itemize}

\section{Related work}
\subsection{Zero-shot learning}
Lampert et al.~\cite{Lampert14DAP} proposed a pioneering two-stage method for ZSL, where a probabilistic classifier was used to predict attribute probabilities and then a Bayesian classifier was used to classify the image based on the attribute probabilities. To achieve end-to-end learning, Frome et al.~\cite{Frome13DeViSE} proposed a visual-to-semantic embedding method where visual features extracted by a CNN were projected into the semantic feature space by a linear function which was trained by a compatibility loss. Based on this work, lots of methods have been proposed to improve the visual-semantic mapping by adopting different loss functions or different mapping functions~\cite{akata2015sje,Akata16ALE,Yu2018TransductiveZL,xian2016LATEM,Zhang2020DeepTN,Zhang2020TowardsED,Pan2020TowardsZL,Luo2020AND,Hou2020DiscriminativeCC}. For instance, ALE~\cite{Akata16ALE} employed a bilinear compatibility function to embed the visual features and semantic features. ESZSL~\cite{romera2015ESZSL} used a nonlinear neural network to map the visual features. LATEM~\cite{xian2016LATEM} embedded the visual features with a piecewise linear function which was trained by a ranking based objective function. To alleviate the hubness problem~\cite{zhang2017DEM} in the visual-to-semantic embedding methods, some works~\cite{Changpinyo17EXEM,Geng2020GuidedCF,Zhang2020APZ,Luo2018ZeroShotLV} proposed to learn a semantic-to-visual embedding, where they mapped semantic features into visual features by a regression function. DEM~\cite{zhang2017DEM} used a two-layer neural network to learn a discriminative visual features space from the semantic feature space. EXEM~\cite{Changpinyo17EXEM} employed SVR to predict the visual exemplar of each class from the corresponding semantic feature. All the aforementioned methods (named as the embedding based method) focused on learning a discriminative embedding space to establish the mapping between visual and semantic features. As another research direction, the generative methods~\cite{Kim2020UnseenIG,liu2020APNet,keshari2020OCDZSL,Vyas2020LrGAN,Ma2020SimilarityPF,Gao2020ZeroVAEGANGU,Long2018ZeroShotLU} have received increasing attention recently due to their superior performances in GZSL. All the generative methods used a conditional generator to generate many unseen-class samples conditioned on their corresponding semantic features and then trained a classifier using these fake unseen-class samples with corresponding labels to classify real unseen-class ones. Their differences mainly lie in the adopted generator and the way to regularize the generator. f-CLSWGAN~\cite{Xian18FCLSWGAN} used a generative adversarial network (GAN) to generate visual features and a pre-trained classifier was used to regularize the feature generation. DASCN~\cite{NiZ019dascn} employed a dual GAN architecture to train a visual feature generator and a semantic feature generator in a cycle manner. CVAE~\cite{Mishra2018CVAE} employed a conditional variational autoencoder (VAE) to generate visual features. CADA-VAE~\cite{schonfeld2019CADA-VAE} used two VAEs to map the visual features and semantic features into the common latent space and generated unseen-class latent features to train an unseen-class classifier. Since an arbitrary number of fake unseen-class samples could be generated by the generative methods, the generative methods could effectively alleviate the bias problem in the GZSL tasks, significantly improving the GZSL performance against the embedding based methods. However, both the generative methods and the embedding based methods generally suffer from the inconsistency between visual features and semantic features, since they employ global visual features extracted by a pre-trained CNN which lack of some local visual information with fine-grained semantics . More recently, some works~\cite{elhoseiny2017ZSLPP,Li18ldf,Zhu19SGML,Xie_2019_AREN,liu2019LFGAA,Jia2020DeepUE} aimed to improve the consistency between visual features and semantic features by employing a fine-tuned CNN with novel architecture to extract more discriminative and semantically consistent visual features, which we denote by fine-tuned feature based methods.

\subsection{Domain-aware generalized ZSL}
In the generalized ZSL, most existing methods suffer from a recognition bias towards the seen-class domain due to the fact that the testing unseen-class data are out-of-distribution (OOD) data compared to the seen-class data. Recently, OOD detection~\cite{Hendrycks17Baseline,Liang18Enhancing,Hendrycks19OE} has received much attention in deep learning community. Inspired by the OOD detection, several works~\cite{socher2013CMT,mandal2019CE-WGAN-OD,dvbe2020} have introduced the OOD detection techniques to alleviate the bias problem. Socher et al.~\cite{socher2013CMT} used a Gaussian mixture model to estimate the probability to be an outlier for each input and detected out the outliers with a threshold. Mandal et al.~\cite{mandal2019CE-WGAN-OD} used a conditional GAN to generate fake OOD data (i.e. fake unseen-class data) and trained a classifier which makes high-entropy predictions on unseen-class data to differentiate the seen and unseen class data. Min et al.~\cite{dvbe2020} proposed an adaptive margin second-order embedding model to minimize the entropies of predictions on seen-class data and then differentiated the seen-class data and unseen-class data according to their entropies. Besides, some other domain-aware modules~\cite{ChaoCGS16,Liu2018DCN,atzmon2019COSMO} were also proposed to tackle the bias problem. Chao et al.~\cite{ChaoCGS16} proposed a simple calibrated stacking technique which subtracts a constant from the probabilities of seen classes. Liu et al.~\cite{Liu2018DCN} proposed a calibration network to reduce the confidences of the seen-class data on unseen classes. Atzmon et al.~\cite{atzmon2019COSMO} computed the probabilities to be seen/unseen class for each input instead of making a hard decision and combined the seen/unseen probabilities with the class-level posterior probabilities to finally classify the input.

\section{Semantic-diversity transfer network}
Before elaborating the proposed method, we first introduce the definition of ZSL. Suppose a labeled seen-class dataset $\mathcal{D}_{S} = \{(x_{n}, y_{n}) \mid n = 1, 2, \cdots, N\}$ and a semantic feature set $\mathcal{E} = \{e_{y} \mid y \in Y \}$ is given in the training stage, where $x_{n} \in \mathbf{R}^{V}$ is the $n$-th labeled visual feature, $y_{n}$ is the class label of $x_{n}$, which belongs to the seen-class label set $Y^{s}$, $N$ is the number of visual features in $\mathcal{D}_{S}$, and $e_{y} \in \mathbf{R}^{S}$ is the semantic feature of the class $y$ in the total class label set $Y$, which not only includes the seen-class label set $Y^{s}$, but also includes the unseen-class label set $Y^{u}$. Note that the unseen-class label set $Y^{u}$ is disjoint with the seen-class label set $Y^{s}$. In the testing stage, given a testing visual feature set $X$, in the conventional ZSL setting, the task is to learn a mapping $f: X \to Y^{u}$ from the test set $X$ to the unseen-class label set $Y^{u}$. In the generalized ZSL setting, the task is to learn a mapping $f: X \to Y$ from the test set $X$ to the total class label set $Y$.

\begin{figure}[t]
	\centering
	\includegraphics[width=1.2\linewidth]{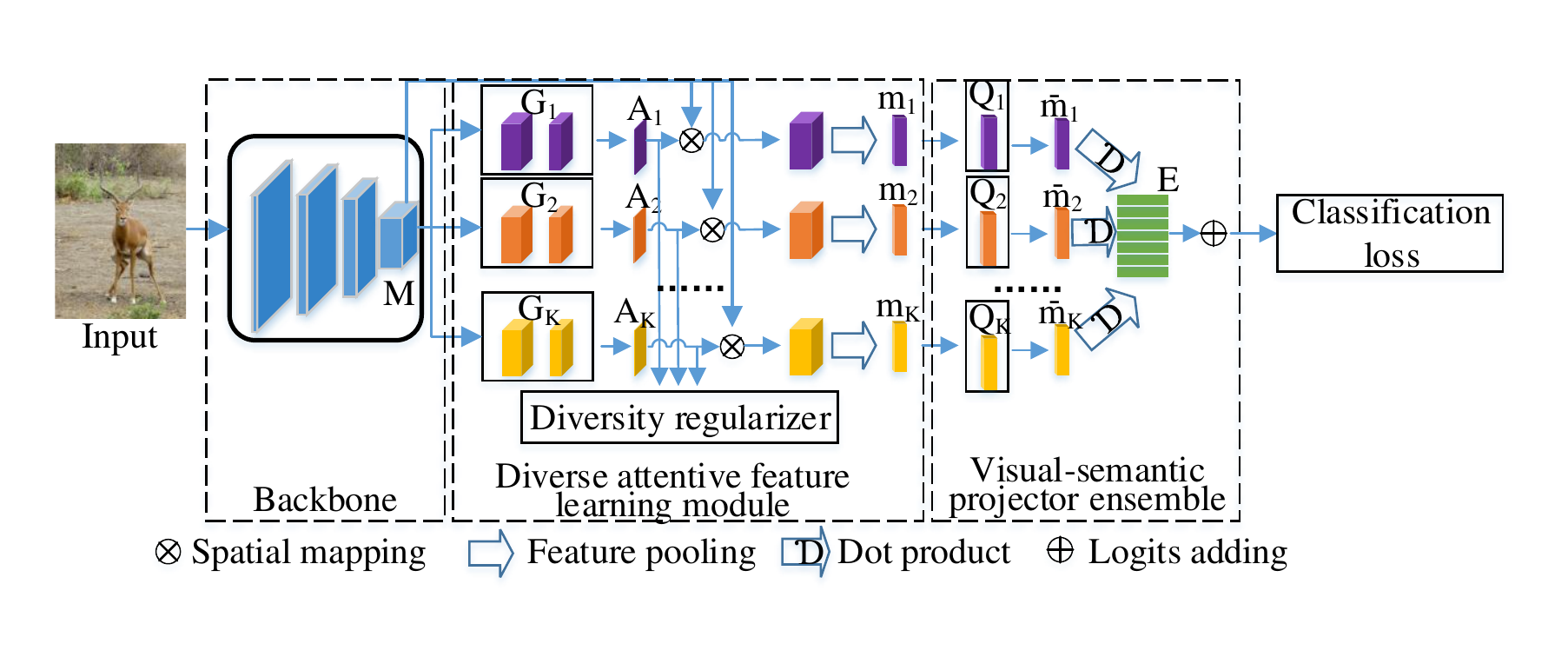}
	\caption{Architecture of the proposed SetNet.}
	\label{fig2}
\end{figure}

In this section, we propose the semantic-diversity transfer network (SetNet) whose architecture is shown in Figure~\ref{fig2}. SetNet consists of three parts: a backbone for extracting general visual features, a diverse attentive feature learning module for extracting semantically diverse local visual features and a visual-semantic projector ensemble for modeling the diverse visual-semantic relations. In the following, we describe the diverse attentive feature learning module and the visual-semantic projector ensemble in detail. In addition, we present a comparison between SetNet and some related works.

\subsection{Diverse attentive feature learning module}
Here our goal is to extract multiple local visual features with semantic diversity. We achieve this goal by firstly extracting multiple local features via a multiple-attention module and then enforcing some semantic diversity between them via a diversity regularizer. Specifically, for a given input image $x$, it is firstly fed into the CNN backbone $F(\cdot)$, generating the general feature maps $M$. Then we compute $K$ spatial attention maps $\{A_{k} \mid k=1, 2, \cdots, K\}$ via $K$ convolutional attention modules $\{G_{k}(\cdot) \mid k=1, 2, \cdots, K\}$, each of which localizes a semantically meaningful region. Lastly we obtain $K$ attentive feature vectors $\{m_{k} \in \mathbf{R}^{V} \mid k=1, 2, \cdots, K\}$ by applying the $K$ spatial attention maps to the general feature maps via a spatial mapping $P_{1}(\cdot)$ followed by an average pooling $P_{2}(\cdot)$. The above procedure is formulated as:
\begin{equation}
	m_{k} = P(F(x), G_{k}(F(x))) \quad k=1, 2, \cdots, K \\
	\label{eq1}
\end{equation}
where $P(\cdot)$ is the concatenation of the $P_{1}(\cdot)$ and the $P_{2}(\cdot)$, which are both parameter-free, while the parameters of $F(\cdot)$ and $G_{k}(\cdot)$ will be trained end-to-end under the guidance of a classification metric presented in the next subsection. Note that the general feature maps $M$ which are generated at the last convolutional layer of the CNN backbone contain rich high-level visual information about the inputs, most existing methods apply a global pooling operation to the general feature maps to extract visual features, resulting in the loss of visual information with fine-grained semantic meanings. Conversely, we extract local visual features with semantic meanings by localizing some semantically meaningful regions via the multiple-attention module. Hence, the visual features learned by our method are more consistent with semantic attributes.

Only by the aforementioned multiple-attention operation, the learned local visual features are not guaranteed to be diverse. To explicitly encourage the diversity, we propose a regularizer on these spatial attention maps for making different spatial attention maps to focus on different spatial locations. Specifically, since the spatial attention map $A_{k}$ is a softmax output, meaning that it is a probabilistic distribution, we could measure the statistic divergence between spatial attention maps by the Helliinger distance metric. To compute Helliinger distance, we firstly vectorize the spatial attention map $A_{k} \in \mathbf{R}^{H \times W}$ into $a_{k} \in \mathbf{R}^{T}, T=H*W$. We denote the $t$-th element in $a_{k}$ by $a_{k}^{t}$, then compute Helliinger distance between two vectorized spatial attention maps $a_{i}$ and $a_{j}$ by:
\begin{equation}
	H^{2}(a_{i},a_{j}) = 1 - \sum_{t=1}^\mathbf{T} (\sqrt{a_{i}^{t} a_{j}^{t}})
	\label{eq2}
\end{equation}
where $H^{2}$ is the square of Helliinger distance which is widely used as a proxy to represent the Helliinger distance. Note that other statistical divergence metrics (e.g. KL divergence) could also be used in this method, we simply use the Helliinger distance due to its simplicity and symmetry. Then, we define the diversity regularizer as the sum of square Helliinger distances between pairwise vectorized spatial attention maps as follows:
\begin{equation}
	\begin{split}
	L_{div} &= \sum_{i=1}^K \sum_{j=1,j \ne i}^K H^{2}(a_{i},a_{j})  \\
			&= \sum_{i=1}^K \sum_{j=1,j \ne i}^K H^{2}( \mathcal{V}(G_{i}(F(x))),\mathcal{V}(G_{j}(F(x))))\\
	\end{split}
	\label{eq3}
\end{equation}
where $\mathcal{V}(\cdot)$ is the vectorization operation, and the others are the same as Equation~\ref{eq1}. The diversity regularizer $L_{div}$ is used to measure the divergence between spatial attention maps, i.e. the difference between spatial locations focused by different spatial attention maps. Hence, we could maximize the diversity of local visual features by maximizing the diversity regularizer.

\subsection{Visual-semantic projector ensemble} 
Here our goal is to learn a visual-semantic projector ensemble to model the relations between visual features and semantic features from diverse local perspectives. Suppose $x$ is the input image and $y$ is the class label of $x$. The semantic feature of $y$ is also known, denoted by $e_{y} \in \mathcal{E}_{t}$, where $\mathcal{E}_{t} \in \mathbf{R}^{S \times D}$ is the testing semantic feature set which contains $D$ class-level semantic features and each of them is a $S$-dimensionality vector. In the above subsection, we have obtained $K$ semantically diverse local visual features $\{m_{k} \in \mathbf{R}^{V} \mid k=1, \cdots, K\}$. Here, we project each visual feature $m_{k}$ into the semantic space via a visual-semantic projector $Q_{k}(\cdot)$, obtaining the projected visual feature $\bar{m}_{k} \in \mathbf{R}^{S}$. Then, we employ the dot-product metric to measure the similarities between the projected visual feature $\bar{m}_{k}$ and the set of testing semantic features $\mathcal{E}_{t}$. These similarities represent the consistency between a local visual feature and the semantic features. Since we have $K$ local visual features for each input image, i.e. $K$ projected visual features, we average the similarities between $K$ projected visual features and the semantic features to represent the overall consistency. Considering that the local visual features are diverse and the projectors are different from each other, the $K$ projectors are able to establish a more diverse visual-semantic mapping compared to the one-mapping methods. Lastly, a softmax function followed by a cross-entropy loss function is used to train the projectors for making the projected visual features be most similar to the semantic features of their corresponding labels and less similar with the semantic features of other classes. We formulate this as follows:
\begin{equation}
	L_{cls} =  \mathcal{C}( \mathcal{S}(\frac{1}{K} \sum_{k=1}^K ((Q_{k}(m_{k})) \cdot \mathcal{E}_{t})), y)
	\label{eq4}
\end{equation}
where $\mathcal{S}(\cdot)$ is the softmax function and $\mathcal{C}(\cdot)$ is the cross-entropy loss function. Note that every semantic feature in $\mathcal{E}_{t}$ is normalized into an unit sphere. 
According to Equation~\ref{eq4} and Equation~\ref{eq1}, we reformulate the $L_{cls}$ as follows:
\begin{equation}
	\begin{split}
	L_{cls} = & \mathcal{C}( \mathcal{S}(\frac{1}{K} \sum_{k=1}^K ((Q_{k}(P(F(x),G_{k}(F(x))))) \cdot \mathcal{E}_{t})), y)
	\end{split}
	\label{eq5}
\end{equation}
Finally our overall objective function is:
\begin{equation}
	L_{all} = L_{cls} + \lambda L_{div}
	\label{eq6}
\end{equation}
where $\lambda$ is a hyper-parameter to weight the diversity regularizer.

In the ZSL testing stage, given an arbitrary input image $x$, it will be predicted as follows:
\begin{equation}
	y^{*} = \arg \max \sum_{k=1}^K (Q_{k}(P(F(x),G_{k}(F(x))))) \cdot \mathcal{E}_{Y^{u}}
	\label{eq7}
\end{equation}
where $\mathcal{E}_{Y^{u}}$ includes the semantic features of classes in the unseen-class label set $Y^{u}$ and $y^{*}$ is the predicted class belonging to the unseen-class label set $Y^{u}$.

\subsection{Comparison to related works}
We note that several works~\cite{Li18ldf,Xie_2019_AREN,Zhu19SGML} also applied attention mechanism to ZSL. However, we differ from them in three aspects. Firstly, we devise an attention module with a different architecture from those in~\cite{Li18ldf,Zhu19SGML,Xie_2019_AREN}. Li et al. and Zhu et al.~\cite{Li18ldf,Zhu19SGML} used a hard attention which crops out salient regions from the image and extracted visual features by feeding the resized image regions into CNN again, which needs more expensive computing resources. Xie et al.~\cite{Xie_2019_AREN} also used a soft attention but they learned the attention map with average pooling while our method employs a learnable convolutional module for integrating visual information across channels. Secondly, we propose a regularizer to explicitly enforce the semantic diversity of local visual features. Thirdly, the learned local visual features are projected into semantic features via a projector ensemble where each projector takes a distinct local feature as input to ensure knowledge transfer from different local perspectives.

\section{Inner disagreement based domain detection for GZSL}
In the GZSL task, the bias to the seen-class domain significantly impedes the performance on the unseen-class domain. The cause of the bias problem is that the testing unseen-class data are generally out-of-distribution (OOD) data compared to the training seen-class data (i.e. in-distribution (ID) data). Inspired by the OOD detection, we propose an inner disagreement based domain detection module (ID3M) for GZSL. In the following, we detailedly describe the proposed ID3M and its application in GZSL.

\subsection{ID3M}
Here we introduce the inner disagreement based domain detection module (ID3M) which picks out the unseen-class data before class-level classification in GZSL. Since the unseen-class data are unavailable in the GZSL training stage, direct OOD detection training is impracticable. To overcome this problem, we propose a self-contained training scheme which trains multiple sub-domain detection modules (sub-DDMs) with virtual OOD data. In the self-contained training scheme, we split the seen classes into $I$ subsets uniformly. Each time we consider $I-1$ subsets as the ID classes and the remaining one as the virtual OOD classes, and train a sub-DDM using these ID classes and the virtual OOD classes. According to the data split, we could train $I$ different sub-DDMs. Each sub-DDM is trained to make high-confidence predictions on its corresponding ID class data and uniform-distribution predictions on its OOD class data. This is implemented by minimizing cross-entropy loss on ID class data and minimizing Kullback-Leibler (KL) divergence between the predictions on virtual OOD class data and the uniform distribution as follows:
\begin{equation}
	L_{i}^{DDM} = E_{(x,y) \in D_{i}^{ID}} \mathcal{C}(F_{i}^{D}(x), y) + E_{\grave{x} \in D_{i}^{OOD}} KL(F_{i}^{D}(\grave{x}) || U_{i})
	\label{eq8}
\end{equation}
where $x$ and $y$ are an image and the corresponding label from the $i$-th ID dataset $D_{i}^{ID}$, $\grave{x}$ is an image from the $i$-th virtual OOD dataset $D_{i}^{OOD}$, $F_{i}^{D}(x)$ and $F_{i}^{D}(\grave{x})$ are the predictions made by the $i$-th sub-DDM on $x$ and $\grave{x}$ respectively, and $U_{i}$ is an uniform distribution over the classes belonging to the $i$-th virtual OOD class set.

After training, all the sub-DDMs tend to make high-confidence predictions on their corresponding ID data and low-confidence predictions on their virtual OOD data. At the same time, since all the sub-DDMs have not seen the real unseen-class data, they are prone to make predictions with similar confidence level on them. From the perspective of data, the seen-class data are ID data for $\frac{I-1}{I}$ sub-DDMs and virtual OOD data for $\frac{1}{I}$ sub-DDMs, while the unseen-class data are OOD data for all the sub-DDMs, namely, the sub-DDMs have evident disagreement on the seen-class data while relatively small disagreement on the unseen-class data. Hence, we define an \textbf{inner disagreement criterion} to differentiate the seen and unseen class data. Specifically, for a given input $x$, we first sort the confidence scores predicted by the $I$ sub-DDMs on it by:
\begin{equation}
	[ p_{s^{1}}, p_{s^{2}}, \cdots, p_{s^{I}} ] = sort([ p_{1}, p_{2}, \cdots, p_{I} ])
	\label{eq9}
\end{equation}
where $ p_{i} = \max F_{i}^{D}(x) - En(F_{i}^{D}(x)) $ is the normalized confidence score predicted by the $i$-th sub-DDM on the input $x$ and $En(F_{i}^{D}(x))$ is the entropy of the prediction $F_{i}^{D}(x)$, and $p_{s^{i}}$ is the $i$-th largest confidence score on $x$. Then, we define the \textbf{disagreement degree} $d$ as the confidence distance between the mean of the Top $I-1$ largest confidence scores and the smallest confidence score:
\begin{equation}
	d = \frac{1}{I-1} \sum_{t=1}^{I-1} p_{s^{t}} - p_{s^{I}}
	\label{eq10}
\end{equation}
Lastly, we differentiate the seen and unseen class data based on the \textbf{inner disagreement criterion:
if the disagreement degree $d$ on the input is smaller than $\theta$, the input is regarded as an unseen-class data, otherwise a seen-class data.}

We note that some related works~\cite{socher2013CMT,mandal2019CE-WGAN-OD,dvbe2020} also devised an OOD detection module for GZSL, However, we differ from them in that we propose a different method to perform OOD detection. Specifically, our OOD detector is based on the inner disagreement between sub-DDMs which are trained via our proposed self-contained training scheme, while existing works~\cite{socher2013CMT,mandal2019CE-WGAN-OD,dvbe2020} trained single network by maximizing the prediction confidence (or minimizing prediction entropy) on the seen-class data.

\subsection{Classification for GZSL}
Combining the proposed ID3M with SetNet, the flowchart for GZSL is designed as shown in Figure~\ref{fig1}. ID3M firstly picks out the unseen-class data from all the testing data and then the picked unseen-class data are predicted by a SetNet for ZSL (i.e. ZSL-SetNet) according to Equation~\ref{eq7} while the remaining data are predicted by a SetNet for GZSL (i.e. GZSL-SetNet) as follows:
\begin{equation}
y^{*} = \arg \max \sum_{k=1}^K (Q_{k}(P(F(x),G_{k}(F(x))))) \cdot \mathcal{E}_{Y} \\
\end{equation}
where $\mathcal{E}_{Y}$ includes the semantic features of classes in the total class label set $Y$ and $y^{*}$ is the predicted class in the total classes label set $Y$. The others are the same as in Equation~\ref{eq7}. Note that we employ a GZSL model to classify the remaining data instead of a seen-class classifier due to the facts that 1) the GZSL model usually has a considerably high performance on the seen classes, hence using a GZSL model to classify the remaining data has a very small influence on the seen-class accuracy; 2) limited by the accuracy of the OOD detector, the remaining data inevitably includes some real unseen-class data, hence using a GZSL model to classify the remaining data could probably improve the unseen-class accuracy. As for the training of GZSL-SetNet, it is trained in the same way with that of ZSL-SetNet (i.e. same loss function and training data), but note that the parameters of GZSL-SetNet are not necessary to be same with those of ZSL-SetNet. 

\begin{table}[t]
	\caption{Statistics of AWA2, CUB and SUN. Number = the number of samples, Visual-Feat = the visual feature dimensionality, Semantic-Feat = the semantic feature dimensionality, All = the number of all classes, Seen = the number of seen classes, Unseen = the nubmer of unseen classes.}
	\centering
	\resizebox{0.9\columnwidth}{!}{
	\begin{tabular}{ccccccccc}
		\hline
		Dataset&  Number&  Visual-Feat&  Semantic-Feat&  All&  \multicolumn{2}{c}{PS}&  \multicolumn{2}{c}{SS} \\
		\cline{6-7} \cline{8-9}
		& & & & & Seen& Unseen& Seen& Unseen \\ 
		\hline
		AWA2&  	 37,322&	  		2048&	  85&        50&   40&	  10&  40&	  10\\
		CUB&  	 11,788&	  		2048&	  312&       200&  150&   50&  150&   50\\
		SUN&  	 14,340&	  		2048&	  102&       717&  645&   72&  645&   72\\
		\hline
	\end{tabular}
	}
	\label{tab1}
\end{table}
\section{Experimental results}

\subsection{Experimental setup}

\subsubsection{Datasets and evaluation protocol}
The proposed method is evaluated on three commonly used datasets whose original images are available for training, including AWA2~\cite{Xian17Comprehensive}, CUB~\cite{WahCUB_200_2011}, SUN~\cite{patterson2012sun}. Specifically, AWA2 is an animal dataset containing 37,322 images from 50 classes, with 85 attributes provided by experts to describe the semantic feature of each class. CUB is a fine-grained bird dataset which contains 11,788 images from 200 species. Each species is annotated by 312 attributes which are usually used as the semantic feature. SUN is a fine-grained scene dataset which includes 14,340 images belonging to 717 classes and each class is annotated by 102 semantic attributes. Note that the number of samples of each class in SUN is relatively small, with 20 images per class. The statistics about the three datasets are summarized at Table~\ref{tab1}. 

Seen/unseen data split has a huge influence on the (G)ZSL performance. Proposed Split (PS)~\cite{Xian17Comprehensive} and Standard Split (SS) are both available to the above three datasets. Although SS does not strictly satisfy the ZSL setting due to the fact that some unseen classes in SS actually have been seen by the commonly used ImageNet1000 pre-trained CNNs, it is still a widely used data split to evaluate the ZSL performance. Hence, as done in many previous works, we employ both PS and SS to evaluate the proposed method in the conventional ZSL setting, and employ PS to evaluate the proposed method in the generalized ZSL setting. Under both the PS and SS data split, $\{40,150,645\}$ classes are regarded as the seen classes in AWA2, CUB, and SUN respectively, while the remaining $\{10,50,72\}$ classes are regarded as the unseen classes respectively. The statistics about the two data splits are summarized at Table~\ref{tab1}. Following the existing methods, we evaluate the performance in the conventional ZSL setting by average per-class Top-1 accuracy ($ACC$) on unseen classes. In the generalized ZSL setting, $ACC$ of both seen classes and unseen classes are computed to evaluate their performances. Besides, the harmonic mean $H$ of the two $ACC$ is also computed to evaluate the overall performance by:
\begin{equation}
	H = \frac{2 \times ACC_{seen} \times ACC_{unseen}}{ACC_{seen} + ACC_{unseen}}
	\label{eq11}
\end{equation}
For the evaluation of ID3M, since the task of ID3M in GZSL is to pick out unseen-class data from all the testing data and the unseen-class data are generally regarded as negative samples, we employ the True-Negative Rate (TNR) to evaluate performance when the False-Negative Rate (FNR) is set as a series of values.

\subsubsection{Implementation details and comparative methods}
The ImageNet1000 pre-trained ResNet101~\cite{He16resnet} is employed as the backbone (i.e. $F(\cdot)$) of SetNet. The $K$ attention modules $\{G_{k}(\cdot)\}_{1}^{K}$ are jointly implemented by a two-layer convolutional module, whose input-channel number is $2048$ and output-channel number is $K$, and the hidden-channel number is $1024$ for all the three datasets. The spatial mapping and pooling operation $P(\cdot)$ is parameter-free. Each visual-semantic projector $Q_{k}(\cdot)$ in the projector ensemble is implemented by a fully-connected layer whose input-unit number is $2048$ and output-unit number is the dimensionality of corresponding semantic features in the three datasets. In ID3M, the ImageNet1000 pre-trained ResNet101 is also employed as the backbone of the sub-DDMs. The pre-trained ResNet101 takes $224*224$ images as inputs. In the training stage, the inputs are preprocessed by `randomcrop' and `randomflip' operations. In the testing stage, only `centercrop' operation is used for preprocessing. The SGD optimizer is employed to train all the models. For the SetNet training, all the models are trained with $60$ epoches and the learning rates are set as $\{0.0005,0.0005,0.001\}$ in AWA2, CUB and SUN respectively. $\lambda$ is simply set as $0.2$ for all three datasets. The visual-semantic projector ensemble size is selected from $\{1,2,4,6,8,10\}$. For the ID3M training, the models are trained by $50$ epoches with learning rates as $0.0001$ for all the three datasets. $I$ is simply set as $5$ since it is a common divisor of the seen-class numbers $\{40, 150, 645\}$ of the three dataset. $\theta$ is set according to False-Negative Rate which is selected from $\{0.05,0.07,0.09,0.11,0.13,0.15,0.17,0.19\}$. Note that the unseen-class data are not needed in the process to set $\theta$.

We compare the proposed methods with $30$ state-of-the-art methods, including $9$ generative methods: GAZSL\cite{Zhu18GAZSL}, CVAE~\cite{Mishra2018CVAE}, f-CLSWGAN~\cite{Xian18FCLSWGAN}, LiGAN~\cite{li2019LiGAN}, SABR~\cite{paul2019SABR}, ABP~\cite{zhu2019ABP}, LsrGAN~\cite{Vyas2020LrGAN}, CADA-VAE~\cite{schonfeld2019CADA-VAE}, DASCN~\cite{NiZ019dascn}, $12$ embedding based methods: DEVISE~\cite{Frome13DeViSE}, ALE~\cite{Akata16ALE}, ESZSL~\cite{romera2015ESZSL}, LATEM~\cite{xian2016LATEM}, SAE~\cite{Kodirov17SAE}, ReViSE~\cite{Tsai2017ReViSE}, SSE\cite{zhang2015SSE}, SYNC\cite{Changpinyo16SYNC}, DEM~\cite{zhang2017DEM}, MLSE~\cite{ding2019MLSE}, TCN~\cite{jiang2019TCN}, APNet~\cite{liu2020APNet}, and $9$ fine-tuned feature based methods: RN~\cite{Sung18rn}, QFSL~\cite{Song18QFSL}, SCoRe~\cite{MorgadoV17SCoRe}, SP-AEN~\cite{chen2018SPAEN}, LDF~\cite{Li18ldf}, SGML~\cite{Zhu19SGML}, AREN~\cite{Xie_2019_AREN}, LFGAA~\cite{liu2019LFGAA}, DAZLE~\cite{huynh2020DAZLE}.

\begin{table}[!h]
	\centering
	\caption{Comparative results ($ACC$) in the conventional ZSL setting on AWA2, CUB, and SUN. Generative: generative methods; Embedding: embedding based methods; Fine-tuned: fine-tuned feature based methods.}
	\resizebox{.75\columnwidth}{!}{
		\begin{tabular}{c|c|cc|cc|cc}
			\hline
			& Method& \multicolumn{2}{c}{CUB}& \multicolumn{2}{c}{AWA2}& \multicolumn{2}{c}{SUN} \\
			\cline{3-4} \cline{5-6} \cline{7-8}
			& &  SS&  PS&  SS&  PS&  SS&  PS \\
			\hline
			\multirow{7}{*}{Generative}
			& CVAE\cite{Mishra2018CVAE}&  -&  52.1&  -&  65.8&  -&  61.7 \\
			& GAZSL\cite{Zhu18GAZSL}&  -&  55.8&  -&  70.2&  -&  61.3 \\
			& f-CLSWGAN\cite{Xian18FCLSWGAN}&  -&  57.3&  -&  -&  -&  60.8 \\
			& LiGAN\cite{li2019LiGAN}&  -&  58.8&  -&  -&  -&  61.7 \\
			& SABR\cite{paul2019SABR}&  -&  63.9&  -&  65.2&  -&  62.8 \\
			& ABP\cite{zhu2019ABP}&  -&  58.5&  -&  70.4&  -&  61.5 \\
			& LsrGAN~\cite{Vyas2020LrGAN}& -&  60.3&  -&  66.4&  -&  62.5 \\
			\hline
			\multirow{11}{*}{Embedding}
			& DEVISE\cite{Frome13DeViSE}&  53.2&  52.0&  68.6&  59.7&  57.5&  56.5 \\
			& ALE\cite{Akata16ALE}&  53.2&  54.9&  80.3&  62.5&  59.1&  58.1 \\
			& ESZSL\cite{romera2015ESZSL}&  55.1&  53.9&  75.6&  58.6&  57.3&  54.5 \\
			& SSE\cite{zhang2015SSE}&  43.7&  43.9&  67.5&  61.0&  25.4&  51.5 \\
			& SYNC\cite{Changpinyo16SYNC}&  54.1&  55.6&  71.2&  46.6&  59.1&  56.3 \\
			& LATEM\cite{xian2016LATEM}&  49.4&  49.3&  68.7&  55.8&  56.9&  55.3 \\
			& SAE\cite{Kodirov17SAE}&  33.4&  33.3&  80.7&  54.1&  42.4&  40.3 \\
			& DEM\cite{zhang2017DEM}&  51.8&  51.7&  80.3&  67.1&  -&  61.9 \\
			& MLSE\cite{ding2019MLSE}&  -&  64.2&  -&  67.8&  -&  62.8 \\
			& TCN\cite{jiang2019TCN}& -&  59.5&  -&  \textbf{71.2}&  -&  61.5 \\
			& APNet\cite{liu2020APNet}& -&  57.7&  -&  68.0&  -&  62.3 \\
			\hline
			\multirow{10}{*}{Fine-tuned}
			& RN\cite{Sung18rn}&  -&  55.6&  -&  64.2&  -&  - \\
			& QFSL\cite{Song18QFSL}&  58.5&  58.8&  72.6&  63.5&  58.9&  56.2 \\
			& SCoRe\cite{MorgadoV17SCoRe}&  59.5&  62.7&  82.8&  61.6&  -&  - \\
			& SP-AEN\cite{chen2018SPAEN}&  -&  55.4&  -&  -&  -&  59.2 \\
			& LDF\cite{Li18ldf}&  67.1&  67.5&  83.4&  65.5&  -&  - \\
			& SGML\cite{Zhu19SGML}&  70.5&  71.0&  83.5&  68.8&  -&  - \\
			& AREN\cite{Xie_2019_AREN}&  70.7&  71.8&  86.7&  67.9&  61.7&  60.6 \\
			& LFGAA\cite{liu2019LFGAA}&  67.6&  67.6&  84.3&  68.1&  62.0&  61.5 \\
			& DAZLE\cite{huynh2020DAZLE}& 67.8&  65.9& -& -& -& - \\
			\cline{2-8}
			& SetNet(Ours)&  \textbf{71.0}&  \textbf{74.0}& \textbf{88.7}&  68.0&  \textbf{66.7}&  \textbf{64.2} \\
			\hline
		\end{tabular}
	}
	\label{tab3}
\end{table}
\subsection{Conventional ZSL results}
In the ZSL setting, we first evaluate the proposed SetNet on the three datasets with both PS and SS data split and then compare it with $27$ existing methods as shown in Table~\ref{tab3}. The compared methods are roughly divided into three categories: the generative methods which are based on visual feature generation, the embedding based methods which focus on learning a discriminative embedding space given the semantic features and the visual features extracted by a pre-trained CNN, and the fine-tuned feature based methods which aim to learn more semantically consistent visual features via re-training a visual feature extractor. Results are reported in Table~\ref{tab3} where the results of existing methods are cited from the original papers or public results~\cite{Xian17Comprehensive}. From Table~\ref{tab3}, we can see that SetNet has achieved significant improvements against the existing methods. Specifically, compared with the embedding based methods and the generative methods, the improvements are evident, especially on CUB and SUN. This demonstrates that the local visual features learned by SetNet are more consistent with the semantic features than the global visual features used by the comparative methods. In addition, SetNet outperforms all the fine-tuned feature based methods by a significant margin about $2.2\%$ on CUB with PS, $2.0\%$ on AWA2 with SS, and $4.7\%$ and $2.7\%$ on SUN with SS and PS respectively. These improvements demonstrate that the proposed visual-semantic projector ensemble could learn a more effective visual-semantic mapping than the one-mapping methods by modeling the visual-semantic relations from diverse local perspectives.

\subsection{Generalized ZSL results}
In the GZSL setting, we present two evaluations on the three datasets with PS data split. The first one performs GZSL task without ID3M while the second one with ID3M. The performances are compared with $24$ existing methods listed in Table~\ref{tab4}. In the first evaluation, SetNet achieves better performances than most existing embedding based methods and fine-tuned feature based methods, which demonstrates the effectiveness of the local feature learning and projector ensemble. However, it is obvious that SetNet suffers from recognition bias to seen-class domain like all the embedding based methods and fine-tuned feature based methods. Hence, we adopt the proposed ID3M in the second evaluation to reduce this bias. The corresponding results in Table~\ref{tab4} show us lots of information. Firstly, SetNet with ID3M achieves huge improvements, especially on AWA2 and CUB with improvements about $30.3\%$ and $14.2\%$ respectively. This is caused by the fact that ID3M achieves significantly excellent performances in distinguishing seen-class/unseen-class data on AWA2 and CUB. Secondly, the most recent state-of-the-art methods~\cite{Xie_2019_AREN,huynh2020DAZLE} also adopt an external module to reduce the bias. Compared with them, our method is able to achieve state-of-the-art results on more datasets. For instance, our method has comparable performance with DAZLE+CAL\cite{huynh2020DAZLE} on AWA2 while it outperforms DAZLE+CAL by $8.7\%$ and $2.8\%$ on CUB and SUN respectively. Thirdly, on CUB and AWA2, SetNet with ID3M also outperforms the generative methods which are famous for their superior GZSL performance, while it has inferior performance on SUN. This is because ID3M achieves a plain performance on SUN which has a large number of classes (717 classes). Actually, the generative methods and the OOD-detection based methods adopt two different strategies to alleviate the bias problem in GZSL. The former reduces the bias by re-weighting the seen-class samples and fake unseen-class samples in the overall loss function while the latter achieves it by distinguishing seen-class/unseen-class samples before class-level classification. The compared results show us that the OOD-detection based methods are superior at handling the bias problem in the GZSL setting where the class number is relatively small while the generative methods have superior performances at the setting where the class number is relatively large. Finally, it is noted that the seen-class ACC drops after adding the ID3M. This is because ID3M could not discriminate seen-class samples from unseen-class samples with a $100\%$ accuracy. A subset of seen-class samples are inevitably misclassified as `unseen-class samples' and they would be definitely wrongly classified since they are classified in the unseen-class space. Actually, there is a balance between seen-class and unseen-class accuracies. As noted in Table~\ref{tab4}, the same problem exists at DAZLE+CAL and AREN+CS and the seen-class accuracy at generative methods is also limited due to the increase of unseen-class accuracy.
\begin{table}[!h]
	\centering
	\caption{Comparative results in the generalized ZSL setting on AWA2, CUB, and SUN. U = $ACC$ of unseen classes, S = $ACC$ of seen classes, H = Harmonic mean of unseen-class $ACC$ and seen-class $ACC$. Generative: generative methods; Embedding: embedding based methods; Fine-tuned: fine-tuned feature based methods.}
	\resizebox{0.9\columnwidth}{!}{
		\begin{tabular}{c|c|ccc|ccc|ccc}
			\hline
			& Method& \multicolumn{3}{c}{CUB}& \multicolumn{3}{c}{AWA2}& \multicolumn{3}{c}{SUN} \\
			\cline{3-5} \cline{6-8} \cline{9-11}
			& &  U&  S&  H&  U&  S&  H&  U&  S&  H \\
			\hline
			\multirow{8}{*}{Generative}
			& CVAE\cite{Mishra2018CVAE}& -& -& 34.5& -& -& 51.2& -& -& 26.7 \\
			& f-CLSWGAN\cite{Xian18FCLSWGAN}& 43.7& 57.7& 49.7& -& -& -& 42.6& 36.6& 39.4 \\
			& LiGAN\cite{li2019LiGAN}& -& -& -& -& -& -& 42.9& 37.8& 40.2 \\
			& SABR\cite{paul2019SABR}& 55.0& 58.7& 56.8& 30.3& 93.9& 46.9& 50.7& 35.1& \textbf{41.5} \\
			& ABP\cite{zhu2019ABP}& 47.0& 54.8& 50.6& 55.3& 72.6& 62.6& 45.3& 36.8& 40.6 \\
			& CADA-VAE\cite{schonfeld2019CADA-VAE}& 51.6& 53.5& 52.4& 55.8& 75.0& 63.9& 47.2& 35.7& 40.6 \\
			& DASCN\cite{NiZ019dascn}& 45.9& 59.0& 51.6& -& -& -& 42.4& 38.5& 40.3 \\
			& LsrGAN\cite{Vyas2020LrGAN}& 48.1& 59.1& 53.0& -& -& -& 44.8& 37.7& 40.9 \\
			\hline
			\multirow{10}{*}{Embedding}
			& DEVISE\cite{Frome13DeViSE}& 23.8& 53.0& 32.8& 17.1& 74.7& 27.8& 16.9& 27.4& 20.9 \\
			& ALE\cite{Akata16ALE}& 23.7& 62.8& 34.4& 14.0& 81.8& 23.9& 21.8& 33.1& 26.3 \\
			& ESZSL\cite{romera2015ESZSL}& 12.6& 63.8& 21.0& 5.9& 77.8& 11.0& 11.0& 27.9& 15.8 \\
			& LATEM\cite{xian2016LATEM}& 15.2& 57.3& 24.0& 11.5& 77.3& 20.0& 14.7& 28.8& 19.5 \\
			& SAE\cite{Kodirov17SAE}& 7.8& 54.0& 13.6& 1.1& 82.2& 2.2& 8.8& 18.0& 11.8 \\
			& DEM\cite{zhang2017DEM}& 19.6& 57.9& 29.2& 30.5& 86.4& 45.1& 20.5& 34.3& 25.6 \\
			& ReViSE~\cite{Tsai2017ReViSE}& 37.6& 28.3& 32.3& 46.4& 39.7& 42.8& 24.3& 20.1& 22.0 \\
			& MLSE\cite{ding2019MLSE}& 22.3& 71.6& 34.0& 23.8& 83.2& 37.0& 20.7& 36.4& 26.4 \\
			& TCN\cite{jiang2019TCN}& 52.6& 52.0& 52.3& 61.2& 65.8& 63.4& 31.2& 37.3& 34.0 \\
			& APNet\cite{liu2020APNet}& 48.1& 55.9& 51.7& 54.8& 83.9& 66.4& 35.4& 40.6& 37.8 \\
			\hline
			\multirow{9}{*}{Fine-tuned}
			& QFSL\cite{Song18QFSL}& 33.3& 48.1& 39.4& 52.1& 72.8& 60.7& 30.9& 18.5& 23.1 \\
			& RN\cite{Sung18rn}& 38.1& 61.1& 47.0& 30.0& 93.4& 45.3& -& -& - \\
			& LFGAA\cite{liu2019LFGAA}& 36.2& 80.9& 50.0& 27.0& 93.4& 41.9& 18.5& 40.0& 25.3 \\
			& AREN\cite{Xie_2019_AREN}& 38.9& 78.7& 52.1& 15.6& 92.9& 26.7& 19.0& 38.8& 25.5 \\
			& DAZLE\cite{huynh2020DAZLE}& 42.0& 65.3& 51.1& 25.7& 82.5& 39.2& 21.7& 31.9& 25.8 \\
			& AREN+CS\cite{Xie_2019_AREN}& 63.2& 69.0& 66.0& 54.7& 79.1& 64.7& 40.3& 32.3& 35.9 \\
			& DAZLE+CAL\cite{huynh2020DAZLE}& 56.7& 59.6& 58.1& 60.3& 75.7& 67.1& 52.3& 24.3& 33.2 \\
			\cline{2-11}
			& SetNet(Ours)& 40.0& 76.7& 52.6& 24.6& 89.5& 38.6& 24.4& 37.0& 29.4 \\
			& SetNet+ID3M(Ours)& 64.3& 69.4& \textbf{66.8}& 61.8& 77.9& \textbf{68.9}& 37.7& 34.5& 36.0 \\
			\hline
		\end{tabular}
	}
	\label{tab4}
\end{table}

\subsection{Performance of ID3M}
\begin{figure}[t]
	\centering
	\includegraphics[width=0.9\linewidth]{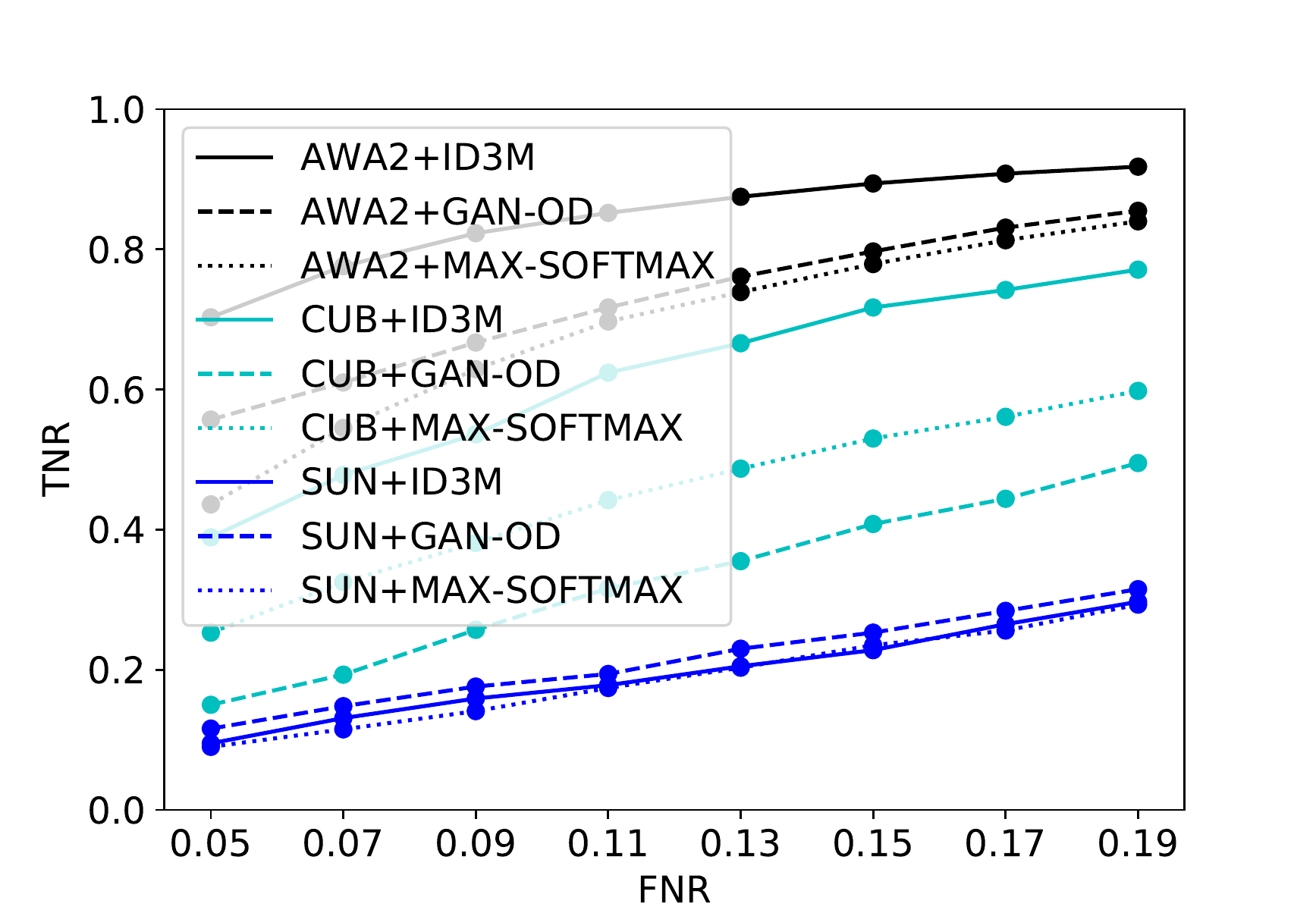}
	\caption{Comparative results (TNR) in the OOD detetion task on AWA2, CUB, and SUN under a series of FNR.}
	\label{fig3}
\end{figure}

We evaluate ID3M by performing OOD detection tasks on AWA2, CUB, and SUN, with FNR selected from $\{0.05,0.07,0.09,0.11,0.13,0.15,0.17,0.19\}$. Since OOD detection methods are usually not evaluated on AWA2, CUB, and SUN, we compare the proposed ID3M with two state-of-the-art OOD detection methods: MAX-SOFTMAX~\cite{Hendrycks17Baseline} and GAN-OD~\cite{mandal2019CE-WGAN-OD}, whose codes are available. The comparative results are shown in Figure~\ref{fig3} where the results of MAX-SOFTMAX and GAN-OD are obtained using the public codes. From Figure~\ref{fig3}, we can see that ID3M achieves significantly superior performances than GAN-OD and MAX-SOFTMAX, especially on CUB and AWA2, which demonstrates the effectiveness of the self-contained training scheme and the inner disagreement criterion. In addition, Figure~\ref{fig3} also shows us that performances of all the three methods on AWA2 and CUB are significantly better than those on SUN. This is because AWA2 and CUB have relatively small numbers of classes (50 and 200 classes respectively), while SUN is a fine-grained dataset with 717 classes and only 20 samples per class. Finally, combining the results in Figure~\ref{fig3} with those in Table~\ref{tab4}, we conclude that an excellent OOD detector can largely improve the GZSL performance, for instance, result from $38.6\%$ to $68.9\%$ on AWA2.

\subsection{Results Analysis}
\subsubsection{Effect of projector ensemble size/attention number}
Here we analyze the effect of ensemble size (i.e. attention number) on the ZSL performance by training SetNet models with different number of visual-semantic projectors. We conduct experiments in the conventional ZSL setting on AWA2, CUB, and SUN with the PS data split. The results are reported on Table~\ref{tab5}, Table~\ref{tab5} shows that the projector ensemble can significantly improve performances compared to single projector (size $= 1$) on all the three datasets. In addition, the performances on all the datasets become insensitive to the ensemble size when the size increases to a certain scale. This is because the diversity of projector's input, i.e. the local visual features will dominate the performances when the projector number reaches a certain value considering the image size is not large.
\begin{table}[t]
	\centering
	\caption{Comparative results ($ACC$) of ZSL-SetNet with different ensemble sizes on AWA2, CUB, and SUN.}
	\resizebox{.65\columnwidth}{!}{
		\begin{tabular}{ccccccc}
			\hline
			Ensemble Size& 1& 2& 4& 6& 8& 10 \\
			\hline
			AWA2& 64.7& 66.6& 65.3& 66.5& 68.0& 67.4  \\
			CUB& 68.9& 69.0& 72.4& 72.6& 74.0& 73.4 \\
			SUN& 60.0& 60.8& 62.2& 63.3& 64.2& 63.8 \\
			\hline
		\end{tabular}
	}
	\label{tab5}
\end{table}

\subsubsection{Effect of disagreement threshold on GZSL performance}
\begin{figure}[t]
	\centering
	\includegraphics[width=0.9\linewidth]{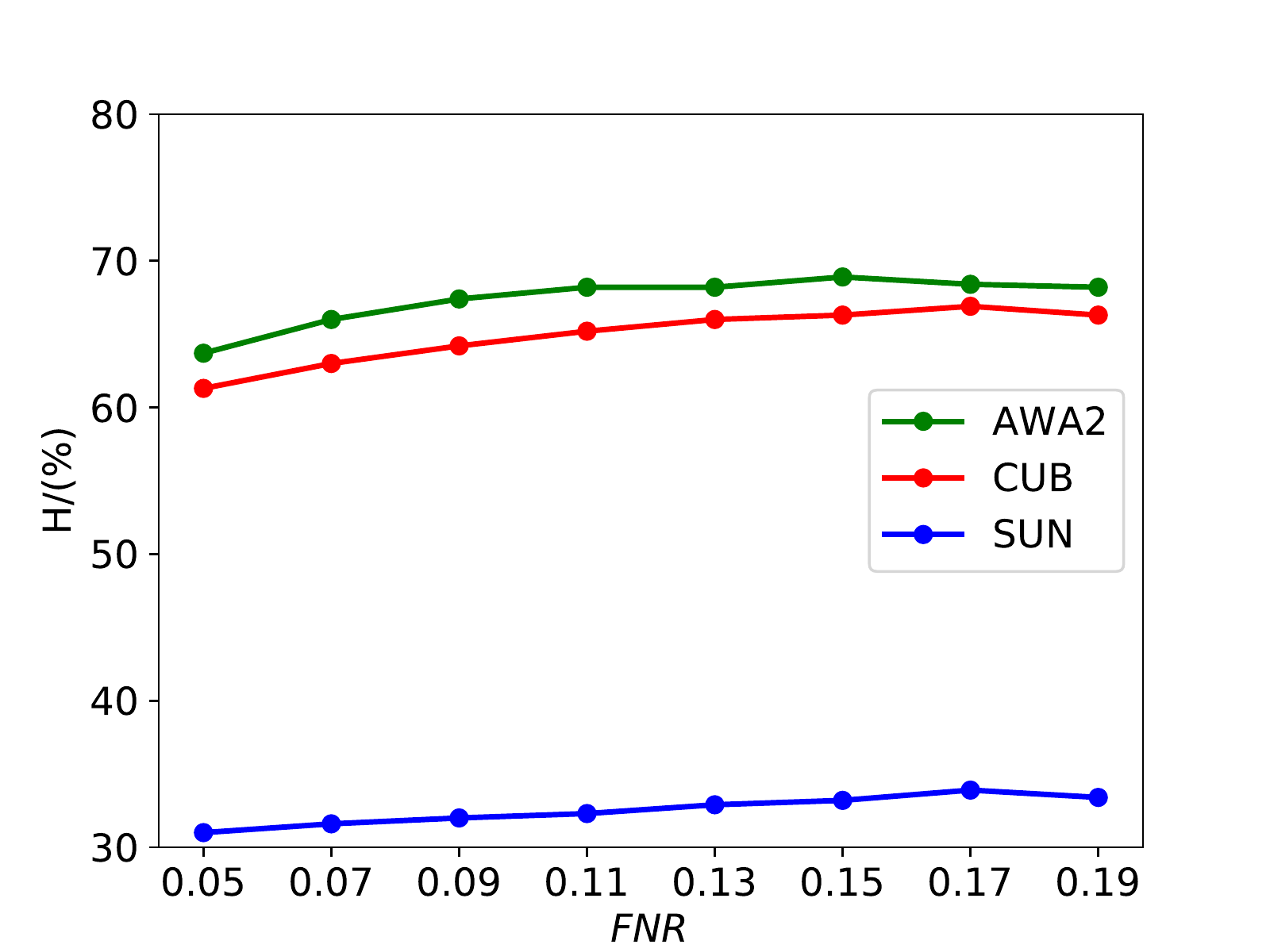}
	\caption{Results with different False-Negative Rates.}
	\label{fig4}
\end{figure}

We propose an inner disagreement based OOD detector (ID3M) for GZSL, where the testing data are firstly classified into `seen class' or `unseen class' according to a set disagreement threshold and then the `seen-class' and `unseen-class' data are classified by a GZSL-SetNet and a ZSL-SetNet respectively. Here we investigate the effect of disagreement threshold ($\theta$) on the GZSL performances by performing GZSL tasks on AWA2, CUB, and SUN under the PS data split setting. $\theta$ is set according to the False-Negative Rate (FNR), which is set as $\{0.05,0.07,0.09,0.11,0.13,0.15,0.17,0.19\}$. The results are shown in Figure~\ref{fig4}. As seen from Figure~\ref{fig4}, at the beginning, the GZSL performances increase as the FNR increases accordingly. This is because when the FNR is larger, more real unseen-class data would be classified as `unseen class' by ID3M and then classified by a specialized ZSL-SetNet, which means a higher unseen-class ACC. Since the harmonic mean (H) of unseen-class ACC and seen-class ACC is mainly influenced by the unseen-class ACC in GZSL, a higher unseen-class ACC results in a higher H. However, exceedingly increasing FNR would make the GZSL performances decrease since it leads to a largely reduced seen-class ACC. On the whole, Figure~\ref{fig4} shows us that the GZSL performances are not very sensitive to the disagreement threshold.

\subsubsection{Sensitiveness to weight of diversity regularizer}
\begin{figure}[t]
	\centering
	\includegraphics[width=0.9\linewidth]{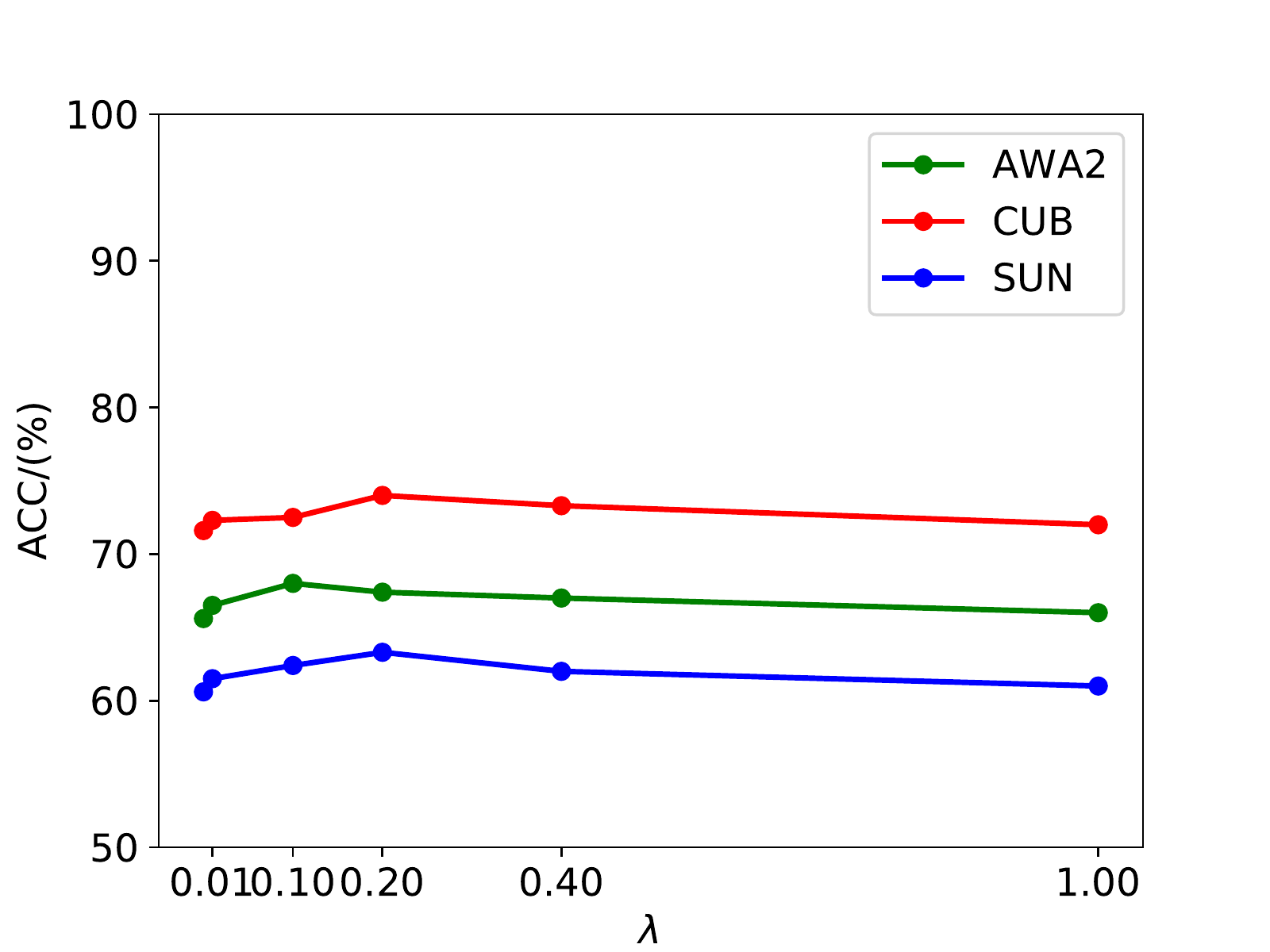}
	\caption{Results with different weights of diversity regularizer.}
	\label{fig5}
\end{figure}

Here we test the sensitiveness of the ZSL performances to the weight of diversity regularizer ($\lambda$) by performing conventional ZSL tasks on AWA2, CUB, and SUN under the PS data split setting with $\lambda = \{0, 0.01, 0.1, 0.2, 0.4, 0.8, 1.0\}$. The results are shown in Figure~\ref{fig5}, where we can see that diversity regularization is beneficial to the ZSL performances and the ZSL performances are not very sensitive to the weight of diversity regularier to a large extent on all the three datasets. However, the ZSL performances would decrease if the weight of diversity regularizer is too large or too small. This is because a quite large $\lambda$ would reduce the influence of the classification objective term on the overall objective and a very small $\lambda$ would ignore the effect of diversity regularizer, both resulting in inferior ZSL performances.

\subsection{Visualization}
\begin{figure}[t]
	\centering
	\includegraphics[width=0.9\linewidth]{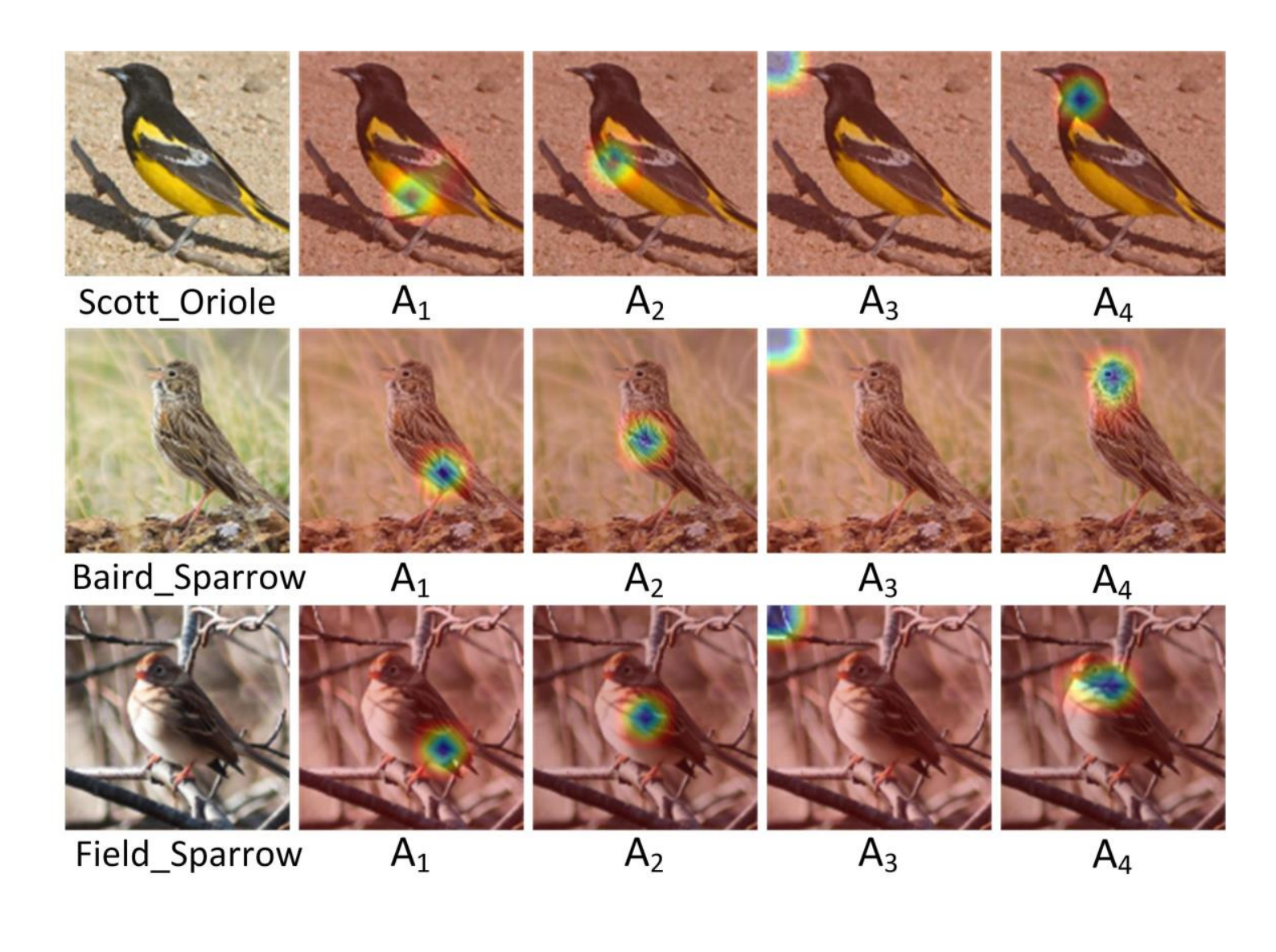}
	\caption{Visualization of semantically salient regions. At each row, the most-left image is the original image, the four images at the right side show the semantically salient regions via heat maps generated by spatial attentions.}
	\label{fig6}
\end{figure}

To show that the multiple-attention module in SetNet has learned semantically meaningful visual features, we visualize the attentive regions generated by SetNet on unseen-class images from CUB. As shown in Figure~\ref{fig6}, different parts like heads, bodies, legs from different bird species are explicitly captured by the multiple-attention module, demonstrating the ability of SetNet to learn local visual features which are consistent with semantic attributes. Besides, it is noted that the region captured by $A_{3}$ seems to have no explicit semantic meaning for the shown images, this is potentially because the visual feature extracted from this region is discriminative for all the images in the whole dataset.

\section{Conclusion and future works}
In this paper, we propose a semantic-diversity transfer network (SetNet) and an inner disagreement based domain detection module (ID3M) to tackle the (G)ZSL problem. SetNet extracts multiple local visual features with semantic diversity via a multiple-attention module and a diversity regularizer, and models the diverse visual-semantic relations by establishing a visual-semantic projector ensemble, facilitating the knowledge transfer from seen classes to unseen classes from diverse local perspectives. ID3M is proposed to alleviate the bias problem in GZSL, which employs a novel self-contained training scheme and detects out unseen-class data based on an inner disagreement criterion. Extensive experimental results show that the proposed method can outperform existing methods with a significant improvement on three public datasets. It is noted from our experimental results that our ID3M performs worse on SUN with a relatively larger number of classes than on the other datasets with a relatively smaller number of classes. In future, we will further investigate how to improve the OOD detection performance on the datasets with a large number of classes.

\section{Acknowledgements}

This work was supported by the National Natural Science Foundation of China (NSFC) under Grants (61991423, U1805264).

\section*{References}

\bibliography{mybibfile}

\end{document}